\documentclass[conference]{IEEEtran}
\usepackage{times}
\usepackage{subcaption}
\usepackage{graphicx}
\usepackage{amsmath}
\usepackage{amssymb}
\usepackage{float}
\usepackage{xcolor}
\newcommand{\textblue}[1]{{\color{blue}#1}}
\newcommand{\textred}[1]{{\color{red}#1}}
\newcommand{\textblack}[1]{{\color{black}#1}}
\definecolor{darkgreen}{RGB}{0, 100, 0}
\newcommand{\textgreen}[1]{{\color{darkgreen}#1}}
\usepackage[numbers]{natbib}
\usepackage{multicol}
\usepackage[bookmarks=true]{hyperref}

\begin{document}

\title{Cross-modality Force and Language Embeddings for Natural Human-Robot Communication}



\author{\authorblockN{Ravi Tejwani}
\authorblockA{Electrical Engineering and\\Computer Science\\
Massachusetts Institute of Technology\\
Cambridge, MA 02139\\
tejwanir@mit.edu}
\and
\authorblockN{Karl Velazquez}
\authorblockA{Electrical Engineering and\\Computer Science\\
Massachusetts Institute of Technology\\
Cambridge, MA 02139\\
kvelaz@mit.edu}
\and
\authorblockN{John Payne}
\authorblockA{Electrical Engineering and\\Computer Science\\
Massachusetts Institute of Technology\\
Cambridge, MA 02139\\
johnpayn@mit.edu}
\and
\authorblockN{Paolo Bonato}
\authorblockA{Department of Physical Medicine, Harvard Medical School\\
Rehabilitation, Spaulding Rehabilitation Hospital\\
Charlestown, MA 02129 USA\\
pbonato@mgh.harvard.edu}
\and
\authorblockN{Harry Asada}
\authorblockA{Mechanical Engineering\\
Massachusetts Institute of Technology\\
Cambridge, MA 02139\\
asada@mit.edu}
}


%

\maketitle

\begin{abstract}
A method for cross-modality embedding of force profile and words is presented for synergistic coordination of verbal and haptic communication. When two people carry a large, heavy object together, they coordinate through verbal communication about the intended movements and physical forces applied to the object. This natural integration of verbal and physical cues enables effective coordination. Similarly, human-robot interaction could achieve this level of coordination by integrating verbal and haptic communication modalities. This paper presents a framework for embedding words and force profiles in a unified manner, so that the two communication modalities can be integrated and coordinated in a way that is effective and synergistic. Here, it will be shown that, although language and physical force profiles are deemed completely different, the two can be embedded in a unified latent space and proximity between the two can be quantified. In this latent space, a force profile and words can a) supplement each other, b) integrate the individual effects, and c) substitute in an exchangeable manner. First, the need for cross-modality embedding is addressed, and the basic architecture and key building block technologies are presented. Methods for data collection and implementation challenges will be addressed, followed by experimental results and discussions.

\end{abstract}

\IEEEpeerreviewmaketitle

\section{Introduction}
\label{sec:Introduction}
As stipulated in Asimov's laws, robots must perform tasks based on human instruction. A central challenge in robotics has been developing effective ways for humans to communicate and give instructions to robots. 

\begin{figure}[h]
    \centering
    \includegraphics[width=1\linewidth]{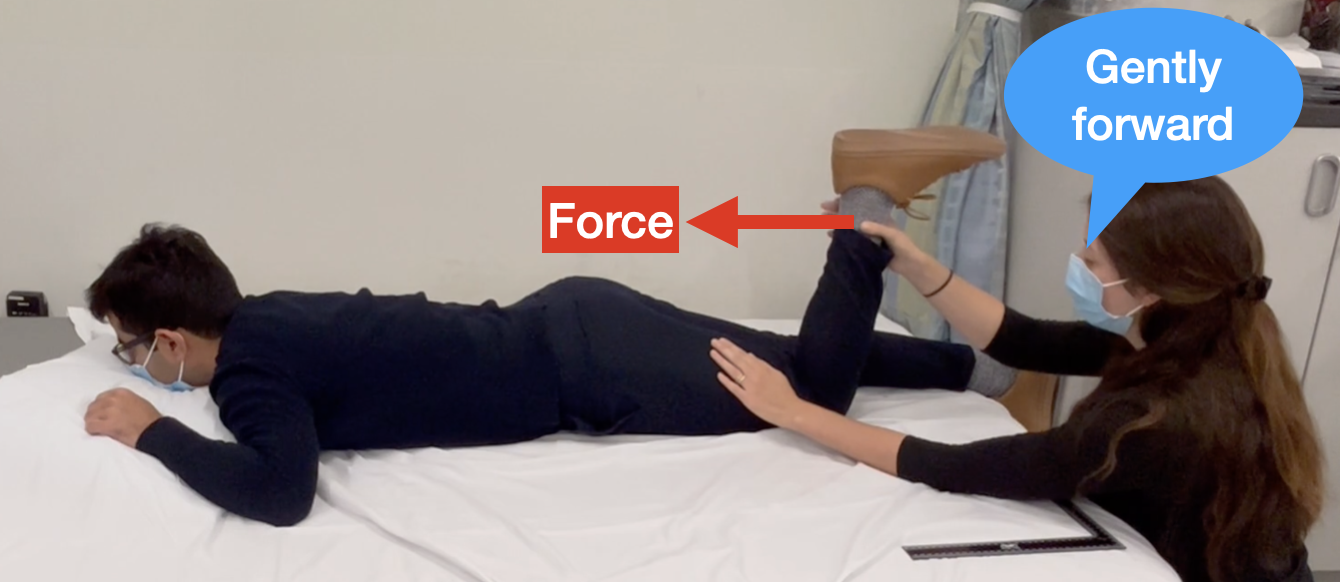}
    \caption{
    Physical therapist from Spaulding Rehabilitation Hospital is seen demonstrating 'hamstring curl' therapy on the patient with neurological injuries 
    \protect\footnotemark . She is instructing a patient to move 'gently forward' while providing an assistive force. }
    \label{fig:therapist-patient}
\end{figure}

Early roboticists attempted to develop robot languages, e.g. VAL \cite{mcgraw1982val}, to describe desired robot actions and tasks. However, despite some success, it became apparent that certain motions and behaviors are difficult, inefficient, or impossible to describe through language alone. This limitation is particularly evident in tasks requiring environmental contact and force/compliance control, where forces and moments are not directly visible. Such manipulative tasks, now termed contact-rich manipulation \cite{pang2023global, beltran2020learning}, often involve subconscious knowledge that humans find difficult to articulate through any form of language. To address this challenge, roboticists developed alternative approaches that bypass the need to translate subconscious skills into language. These approaches include teaching by demonstration \cite{atkeson1997robot}, programming by demonstration \cite{cypher1993watch}, skill acquisition \cite{liu1992transferring} and, more recently, imitation learning \cite{ross2010efficient}.

\footnotetext{Detailed therapy sessions can be seen in our online appendix: \url{https://shared-language-force-embedding.github.io/therapy-sessions/}}

With recent advances in natural language processing, language-grounded robot control has gained significant momentum \cite{tellex2011understanding, matuszek2018grounded}. This approach will play a crucial role in scenarios in which robots and humans interact closely. However, a fundamental question persists: Can language alone effectively convey human intentions, instructions, and desired behaviors to robots?

Consider situations where a human wants a robot to gently touch an object or their body. Such actions are difficult to describe verbally; instead, humans prefer to physically demonstrate the desired gentleness of touch. Yet, physical demonstration alone cannot convey important context, nuance, and reasoning behind the action. This illustrates how language and touch/force are complementary modalities that must be integrated and coordinated for effective human-robot communication.


To study this problem, we collaborated with physical therapists from Spaulding Rehabilitation Hospital who demonstrated various therapeutic techniques. Figure ~\ref{fig:therapist-patient} shows one such demonstration, where the therapist first explains the  procedure verbally and then demonstrates by gently turning the patient's leg. If a robot were to suddenly move the patient's leg, the patient would likely feel uncomfortable or frightened. 
On the other hand, the verbal explanation, ``I will lift your leg gently", is ambiguous to the patient, wondering how gentle is gentle. They may be anxious to see whether it is painful. The therapist starts pushing the leg immediately after giving the brief explanation, demonstrating what she means by lifting the leg gently. This example highlights how language and physical touch/force serve as two distinct but complementary modalities for describing tasks and communicating intended behaviors. The challenge lies in integrating them effectively.  


The goal of the current work is to establish a unified method for representing language and force that facilitates their integration and coordination. We make the following contributions: 
\begin{enumerate}
    \item A framework for cross-modality embedding of force profiles and words \protect\footnotemark, enabling translation between physical force curves and natural language descriptions;
    \item A paired data collection methodology with 10 participants performing language-to-force and force-to-language translations, capturing human intuition about force-language relationships;
    \item Evaluation metrics and results validating the framework's effectiveness and generalization on unseen data;
\end{enumerate}

\footnotetext{Demonstration: \url{https://shared-language-force-embedding.github.io/demo/}}

\section{Related Work}
\label{sec:Related Work}


\subsection{Force-Based Human-Robot Interactions}

Force-based interactions have been studied in the past for human-robot collaborative tasks.
Early work in \cite{asada1989automatic}  demonstrated automatic program generation from force-based teaching data. Furthermore, \cite{kazerooni1993human} established the significance of force feedback in human-robot interfaces, putting down the groundwork for new interaction paradigms. Recent work has significantly improved our understanding of force-based manipulation, with \cite{holladay2019force} demonstrating planning for tool use under force constraints and \cite{holladay2024robust} further extending this to robust multi-stage manipulation tasks.

Research on using force sensing for improved physical human-robot interaction was explored in \cite{haddadin2017robot}  showing methods for learning from demonstration using force data \cite{lee2015learning} and explaining human intent from contact forces \cite{peternel2018robot}. However, these works are applicable to tasks under specific conditions; broader task variability, diverse conditions and contexts, and subtle nuance that language can describe are not considered. 


\subsection{Grounding Natural Language in Robot Actions} 
Natural language has been investigated in literature for grounding language phrases to robot actions and behaviors.  \cite{tellex2011understanding} developed probabilistic approaches for mapping natural language instructions to robot trajectories. 
Building on this, \cite{matuszek2018grounded} showed methods for learning semantic parsers that ground language to robot control policies. Recent work has shown the use of large language models to improve language understanding for robotics \cite{andreas2014grounding} \cite{lynch2020language} . While these approaches map language to robot actions, the tasks are mostly pick-and-place, and more complex manipulative tasks that involve contact forces are excluded. 

\subsection{Multimodal Embeddings in Robotics}
Research in learning shared embedding spaces between different modalities for robotic learning has been explored in the past. \cite{lee2019making} developed cross-modal embeddings between visual and force data for manipulation tasks. \cite{zhang2019neural} showed learning joint embeddings of language and vision for a robot instruction navigation task. 
Although, these approaches have demonstrated the potential of multimodal embeddings in robotics, none have specifically addressed the challenge of creating shared embeddings between force trajectories and natural language descriptions. 

The current work aims to fill this gap by developing and providing a framework of bidirectional translation between physical forces and their linguistic descriptions. Inspired by physical therapists' interactions with patients, we will address the needs for unified representation of language and force profiles and effectiveness of force-language cross-modality embedding to better understand how these strikingly distinct modalities can be integrated.

\section{Preliminaries}
\label{sec:Preliminaries}

\subsection{Coordinate System}
\label{sec:Preliminaries:CoordinateSystem}

We first introduce a coordinate system to consistently define spatial directions and interpret force profiles (Table \ref{tab:direction_axis_mapping}). Each force measurement is a vector $\vec{F}(t)\in\mathbb{R}^3$ with components $(F_x(t),F_y(t),F_z(t))$. Fig. \ref{tab:direction_axis_mapping} shows linguistic direction to its corresponding axis:

\begin{figure}[H]
    \centering
    \includegraphics[width=0.60\linewidth]{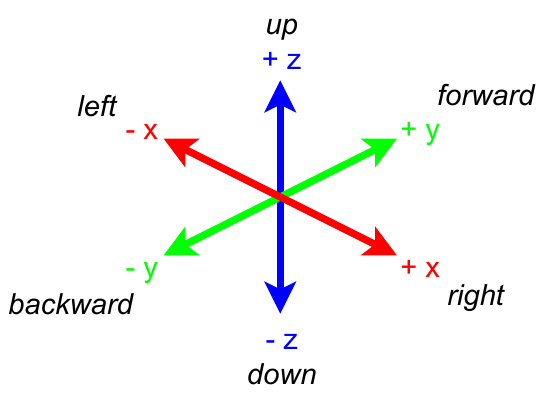}
    \caption{Coordinate system mapping direction words to spatial axes for interpreting force profiles.}
    \label{tab:direction_axis_mapping}
\end{figure}


\subsection{Force Profile}
\label{sec:Preliminaries:Force}
\begin{figure*}[t]
    \centering
    \begin{subfigure}{0.49\linewidth}
        \centering
        \includegraphics[width=0.8\linewidth]{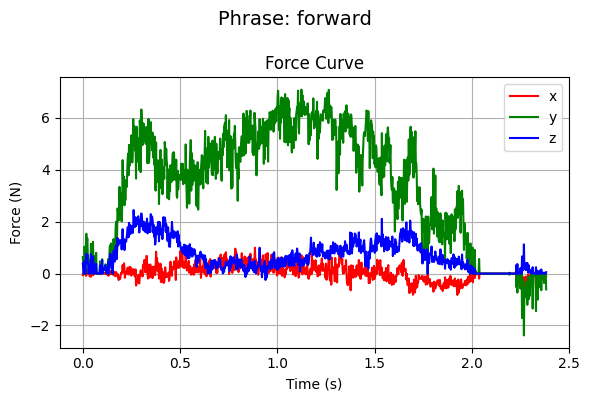} 
        \caption{}
        \label{fig:example_force_profile_a}
    \end{subfigure}
    \begin{subfigure}{0.49\linewidth}
        \centering
        \includegraphics[width=0.8\linewidth]{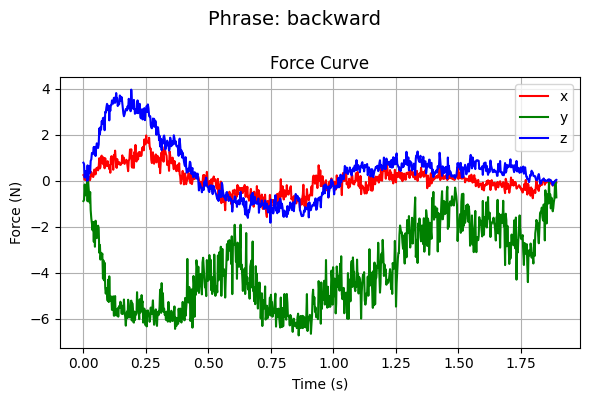} 
        \caption{}
        \label{fig:example_force_profile_b}
    \end{subfigure}
    
    \vspace{2em}  
    
    \begin{subfigure}{0.49\linewidth}
        \centering
        \includegraphics[width=0.8\linewidth]{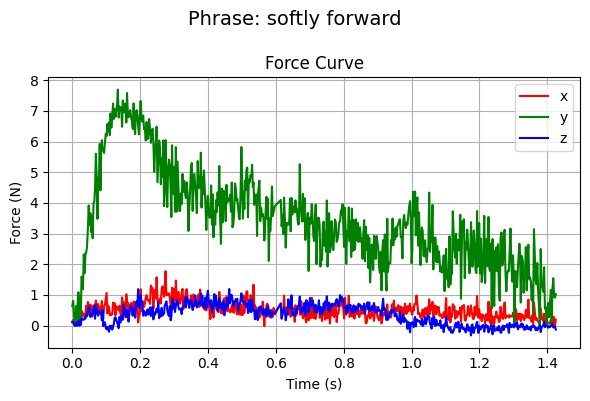} 
        \caption{}
        \label{fig:example_force_profile_c}
    \end{subfigure}
    \begin{subfigure}{0.49\linewidth}
        \centering
        \includegraphics[width=0.8\linewidth]{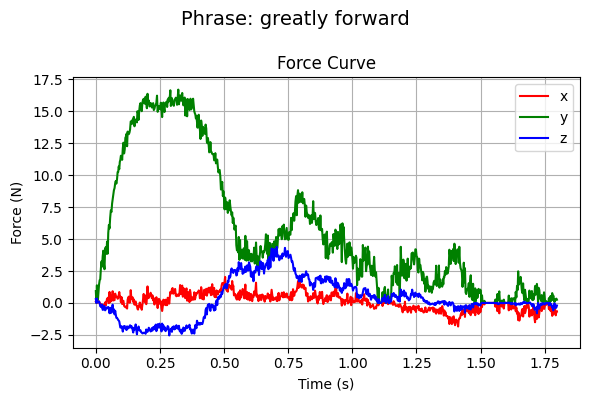} 
        \caption{}
        \label{fig:example_force_profile_d}
    \end{subfigure}
    
    \caption{Examples of corresponding force profiles and phrase pairs. (a,b) Basic motions in forward and backward directions, showing dominant positive and negative y-components respectively. (c,d) Effect of adding modifiers ('softly' and 'greatly') to forward motion, demonstrating how they alter force magnitude while maintaining direction.}
    \label{fig:force_profile_examples}
\end{figure*}

We record time-varying force data using a force-torque sensor mounted on the UR robot's end-effector. For a recording $T$ seconds, we store each timestamp $t_i\in[0,T]$ along with the measured force vector $\vec{F}(t_i)$. We refer to the set of samples, ordered chronologically as a \textit{force profile}. Formally, a force profile is represented as a $4\times\mathcal{N}$ tensor:

\begin{equation}
    \text{Force Profile}=
    \begin{bmatrix}
    t_0 & t_1 & \dots & t_{\mathcal{N}-1} \\
    F_x(t_0) & F_x(t_1) & \dots & F_x(t_{\mathcal{N}-1}) \\
    F_y(t_0) & F_y(t_1) & \dots & F_y(t_{\mathcal{N}-1}) \\
    F_z(t_0) & F_z(t_1) & \dots & F_z(t_{\mathcal{N}-1})
    \end{bmatrix}
\end{equation}

where $t_0=0$ and $t_{\mathcal{N}-1}=T$. Figure \ref{fig:force_profile_examples} shows examples of force profiles, paired with textual instructions.

To interpret forces quantitatively, we adopt Newton's second law of motion:

\begin{equation}
   \vec{p}(t)-\vec{p}(0)=\int_0^T\vec{F}(t)dt=\vec{J}(t)
\end{equation}

where at time $t$, $\vec{p}(t)$ is the momentum, $\vec{F}(t)$ is the applied force, and $\vec{J}(t)$ denotes impulse. This unlocks an intuition of the elementary pillars that describe force profiles: \textit{direction} $\hat{F}(t)$, \textit{magnitude} $||\vec{F}(t)||$, and \textit{duration} $T$.

\subsection{Language}
\label{sec:Preliminaries:Language}

Throughout this paper, we define a \textit{phrase} as an ordered list of words describing a motion or force profile (e.g. ``\textit{slowly forward}", ``\textit{quickly right and up}"). To handle these numerically, we consider two distinct vocabularies.

\subsubsection{Minimal Viable Vocabulary} 
\label{sec:Preliminaries:Language:Binary}
This vocabulary contains 18 direction words (e.g. \textit{left}, \textit{right}, \textit{forward-down}) and 12 modifier words (e.g. \textit{slowly}, \textit{quickly}, \textit{harshly}) that describe variations in force magnitude and duration \cite{irie2021examining}.

\begin{table}[H]
    \centering
    \begin{tabular}{|c|c|}
        \hline
        \textbf{Direction} & \textbf{Modifier} \\
        \hline
        backward & slightly \\
        backward-down & greatly \\
        backward-left & smoothly \\
        backward-right & sharply \\
        backward-up & slowly \\
        down & quickly \\
        down-forward & lightly \\
        down-left & significantly \\
        down-right & softly \\
        forward & harshly \\
        forward-left & gradually \\
        forward-right & immediately \\
        forward-up & \\
        left & \\
        left-up & \\
        right & \\
        right-up & \\
        up & \\
        \hline
    \end{tabular}
    \caption{The minimal viable vocabulary. Direction words describe the overall direction of the force profile, while modifier words describe the magnitude and duration.}
    \label{tab:vocabulary}
\end{table}


\textbf{Binary Phrase Vectors:} The direction words require 18 dimensions and modifier words require 12 dimensions (as shown in Table \ref{tab:vocabulary}).
Each word is encoded as a 31-dimensional basis vector, where the additional dimension represents an empty or null word. These binary phrase vectors are then concatenated to form a 62-dimensional vector, where exactly two positions contain 1 - one in the first 31 dimensions identifying the modifier (or empty modifier) and one in the second 31 dimensions identifying the direction (or empty direction). This binary encoding scheme ensures consistent representation while allowing for partial or incomplete phrases through the accomodation of empty words.


\subsubsection{Extended SBERT Vocabulary} 
\label{sec:Preliminaries:Language:SBERT}
\begin{figure}
    \centering
    \includegraphics[width=0.75\linewidth]{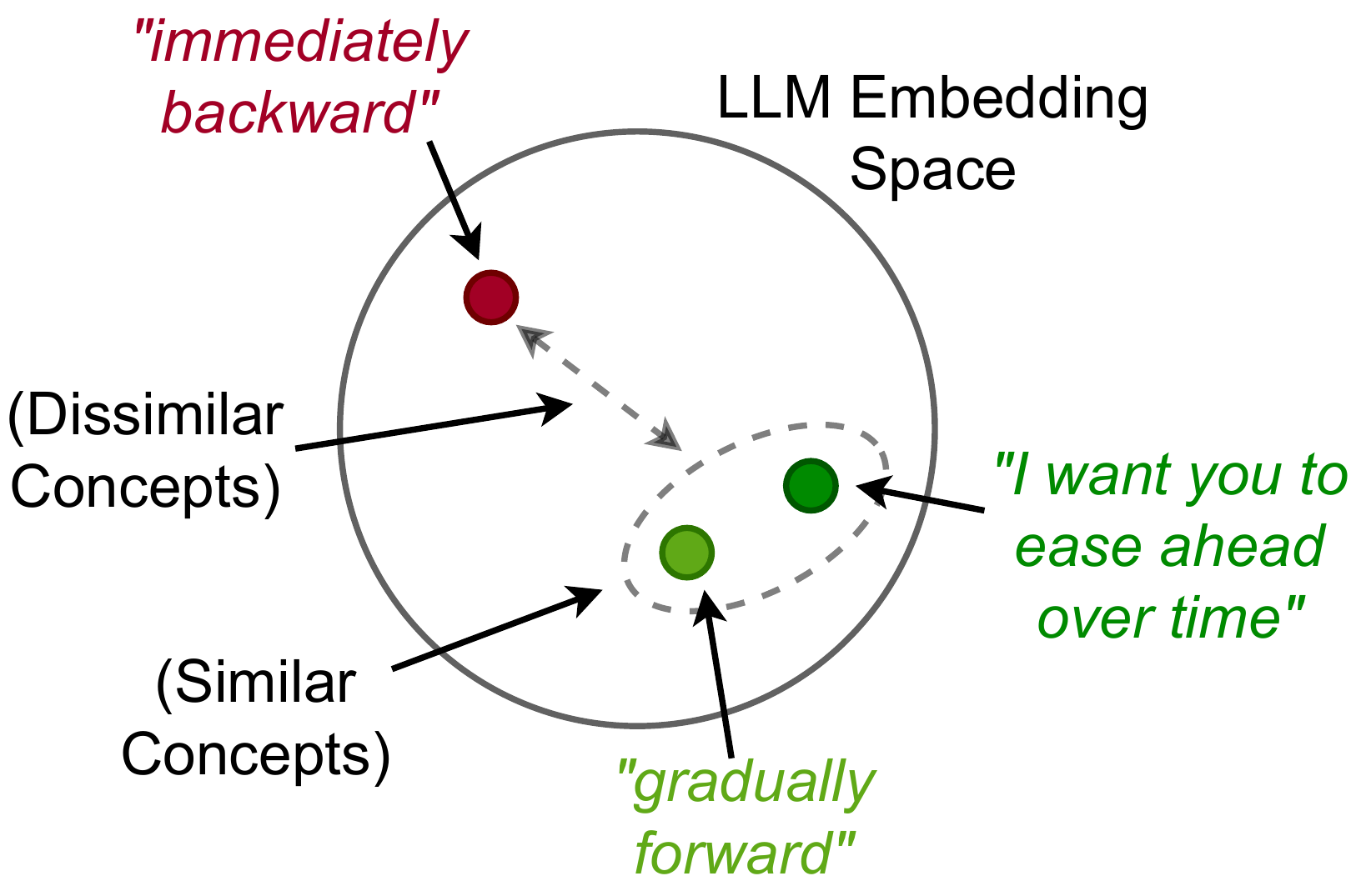}
    \caption{\textblack{Conceptual illustration of the process of matching an arbitrary text input with the most semantically similar phrase using SBERT embeddings. }}\label{fig:nearest_neighbors_method_diagram}
\end{figure}

\textblack{We leverage SBERT (Sentence-BERT) embeddings \cite{reimers2019sentence}, a contextual language model that produces semantically meaningful representations.} 
\textblack{This representation allows for encoding beyond our minimal viable vocabulary, as the SBERT model has been pretrained on diverse textual data. When processing a phrase, we encode the complete text rather than individual words, resulting in a 768-dimensional vector that preserves the semantic meaning of the entire expression (Fig. \ref{fig:nearest_neighbors_method_diagram}). This approach enables our system to handle open-vocabulary inputs while maintaining the shared structure of our force-language embedding space.}

\begin{figure}
    \centering
    \includegraphics[width=1\linewidth]{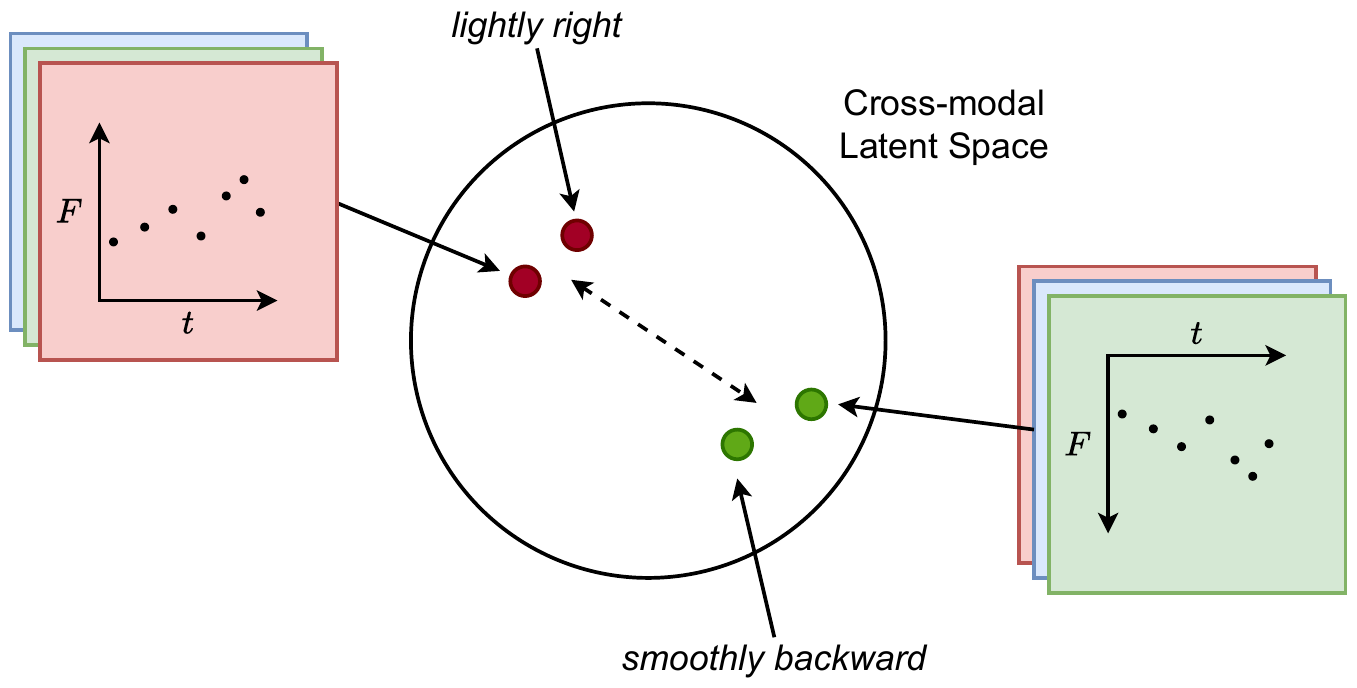}
    \caption{Conceptual illustration of the desired properties of the cross-modality latent space. A pair of corresponding force profile and phrase should be located near each other, measured by a distance metric such as cosine similarity. However, force profiles and phrases that do not correspond should be positioned far away. This demonstrates that similar inputs would be close together and dissimilar inputs would be far apart in the latent space.}
    \label{fig:conceptual_shared_latent_space}
\end{figure}

\subsection{Cross-Modality Embedding (Shared Latent Space)}
\label{sec:Preliminaries:Shared Latent Space}

We learn the cross-modality embedding as a shared latent space $\mathcal{Z}\subset \mathbb{R}^{16}$ to align force profiles and phrases. 
Rather than treating phrases and force profiles as purely distinct modalities, we emphasize on their common representational ground. In this unified latent space, certain force profiles can be naturally described by words (e.g., ``gentle push"), and conversely, phrases can be manifested as force trajectories. 
Specifically, we define encoders $E_{\text{force}}$, $E_{\text{phrase}}$ that map force profiles and phrases, respectively, to a shared embedding $z\in\mathcal{Z}$. We also define decoders $D_{\text{force}}$, $D_{\text{phrase}}$ that map shared embeddings back to forces and phrases, respectively. A contrastive learning objective \cite{hadsell2006dimensionality} encourages embeddings of paired force–phrase data to lie close together in $\mathcal{Z}$ while pushing apart non-matching pairs. This alignment supports:

\begin{itemize}
    \item Force-to-Language Translation: Observed force profiles can be decoded into textual instructions.
    \item Language-to-Force Translation: Written phrases can be transformed into corresponding force trajectories for robotic execution.
\end{itemize}

Together, these capabilities enable more natural interactions in human-robot collaboration by unifying physical force signals and language instructions within a single latent representation \cite{radford2021learning}.

\begin{figure*}
    \centering
    \includegraphics[width=1\linewidth]{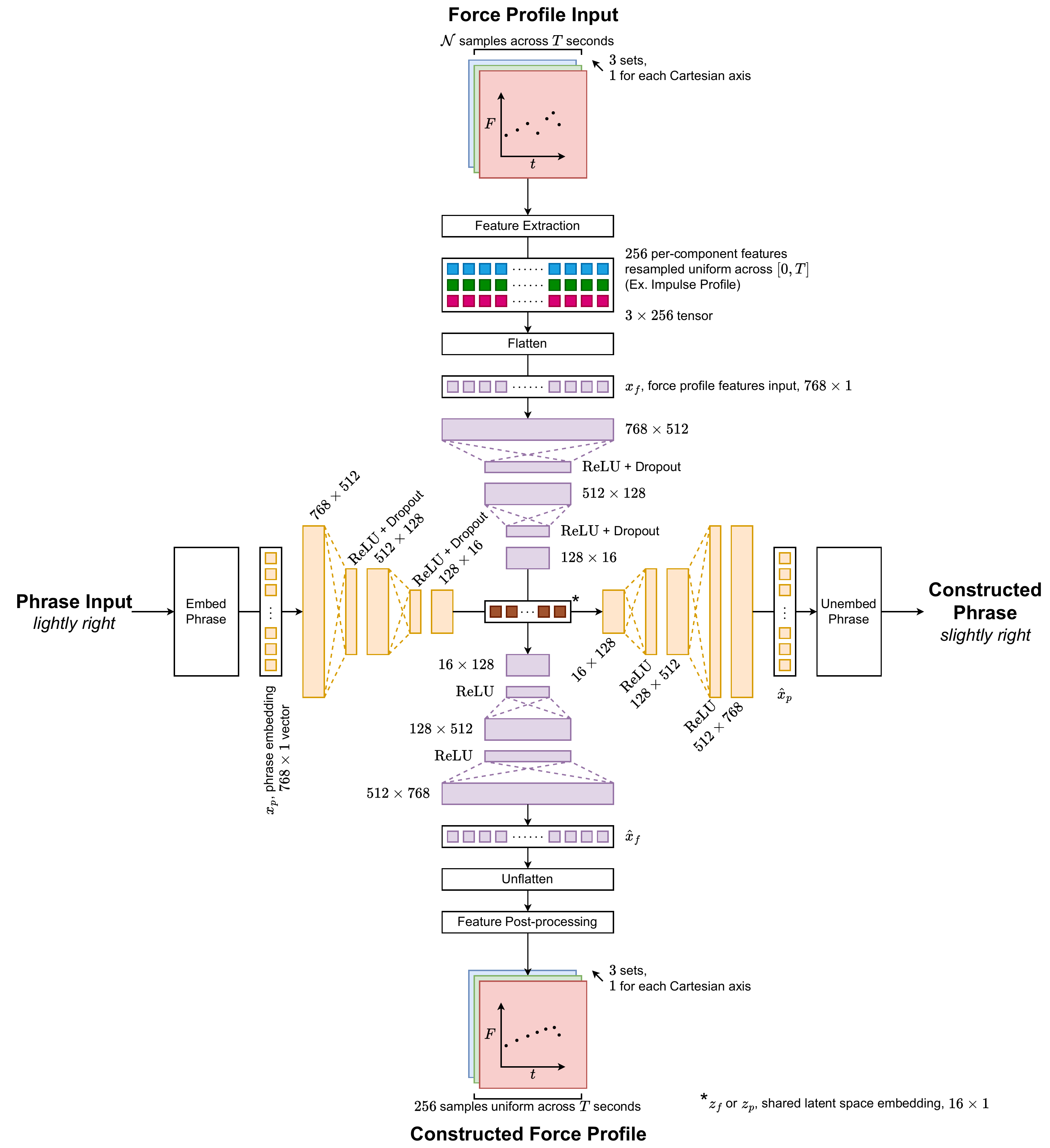}
    \caption{Architecture diagram of the cross-modality  dual autoencoder that represents phrases as a $768\times 1$ embedding generated by S-BERT (\cite{reimers2019sentence}). The phrase embedding input is then passed into the phrase autoencoder, which encodes it into the shared latent space. There the $16\times 1$ embedding can take 2 paths: either be decoded back into a phrase or be translated into a force profile by using the force profile decoder. Force profile inputs are first preprocessed to extract meaningful features for the force profile autoencoder to digest. They are then encoded into the shared latent space. Like phrase inputs, they can either be decoded back into a force profile or be translated into a phrase by using the phrase decoder.}
    \label{fig:architecture_level_1}
\end{figure*}

\section{System Overview}
\label{sec:System Overview}

In this work, we aim to develop a unified representation for physical forces and natural language phrases. Our primary goal is to learn a shared embedding space that allows robots to translate human-applied force profiles into linguistic descriptions and, conversely, generate appropriate force outputs from language instructions. 


We use a UR \cite{ur5} robot for the magnitude, direction, and duration of applied forces over time. To build a dataset that naturally pairs force signals with language, we designed two human-participant procedures. In the Phrase-to-Force procedure (Sec. \ref{sec:Architecture:Data Collection:Phrase-to-Force}), participants receive a brief textual phrase and physically move the robot arm. In the Force-to-Phrase procedure (Sec. \ref{sec:Architecture:Data Collection:Force-to-Phrase}), participants observe an externally applied force trajectory and then describe it in natural language from our minimal viable vocabulary (e.g., “gradually left”). By collecting these paired samples, we obtain a diverse dataset for learning force-language correspondences.

Our proposed model is a dual autoencoder architecture (Fig. \ref{fig:architecture_level_1}), which processes both time-series force profiles and textual phrases. We encode each force profile into a latent representation and similarly embed each phrase into a matching latent space. We train the model with three core objectives that facilitate robust multimodal alignment: (1) reconstruction, which ensures that both forces and phrases can be faithfully recovered from their respective embeddings, (2) contrastive learning, which encourages correct force–phrase pairs to be close in latent space while pushing apart mismatched pairs, and (3) translation, which enables the network to generate a force profile from a given phrase and to describe a given force profile with a textual output.

By optimizing these objectives, the system learns to embed semantically related force and language inputs in close proximity. During inference, a robot can interpret a previously unseen force in linguistic terms or synthesize an appropriate force response for a phrase. We describe the details of the training procedure, architecture, and data preprocessing in Sec. \ref{sec:Architecture}. In subsequent sections, we evaluate how well the learned embeddings capture force-language relationships and demonstrate the system’s capability to perform bidirectional translation between physical forces and natural language.

\section{Architecture}
\label{sec:Architecture}


\subsection{Model}
\label{sec:Architecture:Model}

\subsubsection{Force Profile Input}
\label{sec:Architecture:Model:Force_Profile_Input}

Each raw force profile consists of time-series measurements for the $x$, $y$, and $z$ components of force, recorded at potentially irregular intervals and for varying durations. To create a uniform representation, we first resample each force profile to $\mathcal{N}=256$ evenly spaced time steps spanning a fixed duration $T=4$. This yields a $3\times256$ tensor

\begin{equation}
    \mathbf{F}=
    \begin{bmatrix}
    F_{x,0} & F_{x,1} & \dots & F_{x,255} \\
    F_{y,0} & F_{y,1} & \dots & F_{y,255} \\
    F_{z,0} & F_{z,1} & \dots & F_{z,255}
    \end{bmatrix}
\end{equation}

where $F_{x,i}, F_{y,i}, F_{z,i}$ denote the resampled forces at the $i$-th time step in each axis. Next, we integrate each axis of $\mathbf{F}$ over time to obtain an impulse profile

\begin{equation}
    \mathbf{J}=
    \begin{bmatrix}
    J_{x,0} & J_{x,1} & \dots & J_{x,255} \\
    J_{y,0} & J_{y,1} & \dots & J_{y,255} \\
    J_{z,0} & J_{z,1} & \dots & J_{z,255}
    \end{bmatrix}
\label{eq:impulse_profile}
\end{equation}

where $J_{a,i}=\int_0^{t_i}F_a(t_i)dt$ and $t_i$ is the time associated with the $i$-th sampled resampled point. $\mathbf{J}$ is flattened to form a $768$-dimensional vector $\begin{bmatrix}\mathbf{J}_x&\mathbf{J}_y&\mathbf{J}_z\end{bmatrix}\in\mathbb{R}^{768}$ which serves as input to the force encoder.

\subsubsection{Phrase Input}
\label{sec:Architecture:Model:Phrase Input}
Since neural networks cannot directly process raw text, each phrase must be converted into a fixed-size vector representation that preserves its semantics. We consider two embedding approaches: Binary Phrase Vectors from the Minimal Viable Vocabulary (Section ~\ref{sec:Preliminaries:Language:Binary}), and contextual embeddings from SBERT (Section ~\ref{sec:Preliminaries:Language:SBERT}).

\paragraph{Binary Phrase Vectors} 
In our minimal viable vocabulary, each phrase is composed of exactly two words: a modifier and a direction. Each word is represented as a one-hot encoded vector, with the modifier requiring 31 dimensions and the direction requiring 31 dimensions (including an "empty" position for each). These binary vectors are then concatenated to yield a 62-dimensional phrase representation. This discrete encoding provides a direct mapping between the linguistic components and physical force characteristics.

\paragraph{\textblack{SBERT Embeddings}} 
\textblack{To handle arbitrary verbal instructions beyond our fixed vocabulary, we implement a mapping mechanism using SBERT that identifies the most semantically similar phrase in our minimal viable vocabulary. For any input text $x$, we find the matching MVV phrase $p^*$ using:
\begin{equation}
    p^*=\begin{cases}
        \arg\max_{p\in P}s(E(p),E(x))\;&\text{if }s(E(p),E(x))>\sigma \\
        \text{``"}\;&\text{otherwise}
    \end{cases}
\end{equation}
where $P$ is the set of MVV phrases, $E$ is the SBERT encoder, $s$ is cosine similarity, and $\sigma$ is a threshold parameter. This approach enables our system to accept open-vocabulary input while leveraging our trained force-language mappings.}

\subsubsection{Dual Autoencoders}
\label{sec:Architecture:Model:Dual Autoencoders}

Our framework employs two autoencoders \cite{kramer1991nonlinear} —one for force profiles and one for phrases—to map inputs from distinct modalities into a shared latent space. \textblack{For the phrase autoencoder, we utilize SBERT's 768-dimensional embeddings or the Binary Phrase Vector (62-dimensional embedding) as input, which are then processed through several layers before reaching the 16-dimensional bottleneck layer. This architecture enables the model to compress the semantic information into a compact representation that aligns with force profiles in latent space.}


During training, the encoder learns to capture the essential latent features of the input data, while the decoder learns to reconstruct the input from these latent variables. This unsupervised learning process enables the network to extract a compact representation that generalizes well to novel inputs sharing similar underlying structure.
In our dual autoencoder architecture, both the force and phrase encoders are designed to output latent vectors of the same dimension, ensuring compatibility in the shared cross-modal latent space. This design choice allows us to perform bidirectional translation between force profiles and phrases by using the decoder of one modality on the latent representation produced by the encoder of the other.

Online Appendix \protect\footnotemark details all the specific layers and activation functions and hyperparameters used for each modality. In all cases, layers consist of a linear transformation (i.e., a matrix multiplication) followed by the application of a nonlinear activation function; we use the rectified linear unit (ReLU) for this purpose.

\footnotetext{Online Appendix with framework details: \url{https://shared-language-force-embedding.github.io/framework/}}

\subsection{Multitask Learning}
\label{sec:Architecture:Multitask Learning}

To encourage the model to learn a robust cross-modality latent representation, we adopt a multitask learning strategy that jointly optimizes three related objectives: (1) \emph{reconstruction} of the original inputs from their latent representations, (2) \emph{contrastive} alignment of corresponding force profile and phrase embeddings in the shared latent space, and (3) \emph{translation} between modalities by decoding a latent embedding obtained from one modality into the other. Joint training with these tasks compels the model to extract meaningful latent features that generalize well across modalities while mitigating over-fitting.

Typically, a model is trained by minimizing a single loss function; however, by incorporating multiple loss functions corresponding to related tasks \cite{caruana1997multitask}, our network is guided to form a representation that simultaneously serves several objectives. The overall loss function is a weighted sum of the individual losses:

\begin{equation}
\label{eq:total_loss}
    \mathcal{L}=k_r\mathcal{L}_r + k_z\mathcal{L}_c+k_t\mathcal{L}_t,
\end{equation}

where hyperparameters $k_r,k_z,k_t$ control the relative importance of the reconstruction loss $\mathcal{L}_r$, contrastive loss $\mathcal{L}_c$, and translation loss $\mathcal{L}_t$, respectively. In our experiments, these constants are all set to 1, indicating equal weighting for each task. The loss functions are defined as:

\subsubsection{Reconstruction Loss ($\mathcal{L}_r$) }
\label{sec:Architecture:MultitaskLearning:ReconstructionLoss}

It measures how accurately each autoencoder (force and phrase) reproduces its own input from the latent vector. For force profiles, we use mean squared error; for phrases, the reconstruction metric depends on the chosen representation, with cross-entropy for binary phrase vectors and mean squared error for S-BERT embeddings.

\textblack{For phrase inputs $x_p$ with encoder $E_p$ and decoder $D_p$, the reconstruction loss varies:
\begin{equation}
    \mathcal{L}_{r,p}(x_p)=
    \begin{cases}
        H(x_p,\text{softmax}(D_p(E_p(x_p)))) & \text{binary} \\
        \|x_p-D_p(E_p(x_p))\|^2 & \text{SBERT}
    \end{cases}
\end{equation}
where $H$ is cross-entropy loss.}

\subsubsection{Contrastive Loss ($\mathcal{L}_c$)}
\label{sec:Architecture:MultitaskLearning:ContrastiveLearning}

\begin{figure}
    \centering
    \includegraphics[width=1\linewidth]{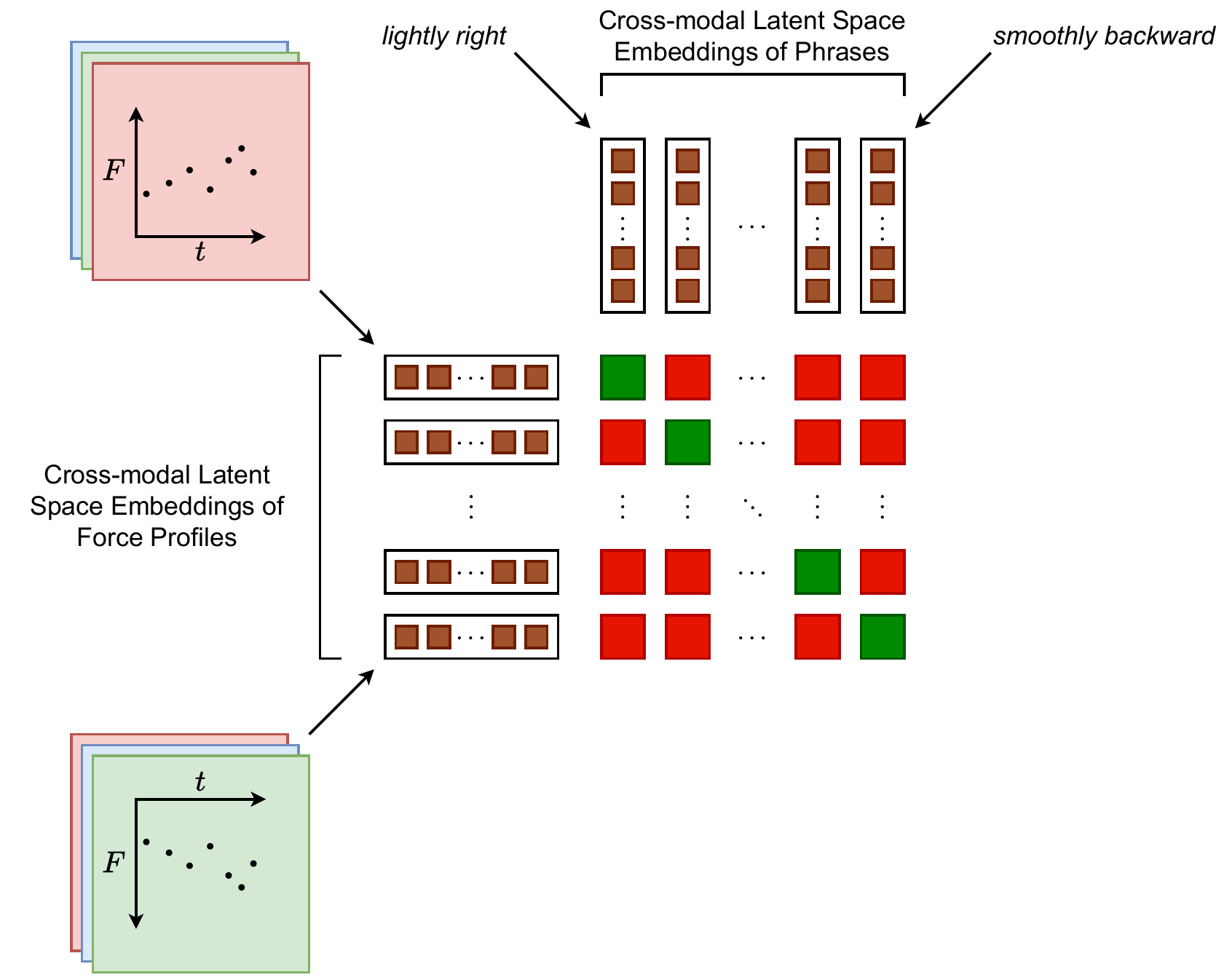}
    \caption{Conceptual illustration of the contrastive loss function. Given a batch of corresponding force profiles and phrases, a distance metric is measured for each pair across modalities. This results in a distance metric matrix. The diagonals refer to the distance metrics of corresponding force profiles and phrases, i.e. positive pairs. Off-diagonal elements refer to negative pairs. The contrastive loss function encourages the distances of positive pairs to be minimized. It also penalizes negative pairs if they start to get too close to each other.}
    \label{fig:contrastive_loss}
\end{figure}

To align the force and phrase embeddings, we employ a contrastive loss \cite{hadsell2006dimensionality} that brings paired (corresponding) latent vectors closer together while pushing apart unpaired (non-corresponding) vectors. This ensures that shared features of matching force and language inputs are learned and represented similarly in the latent space.

\textblack{The contrastive loss for a batch of $n$ force-phrase pairs $(z_f^i, z_p^i)$ in the shared latent space is:
\begin{equation}
    \mathcal{L}_c = \sum_{i=1}^{n}\|z_f^i-z_p^i\|^2 - \lambda\sum_{i=1}^{n}\sum_{j\neq i}^{n}\max(0, m-\|z_f^i-z_p^j\|^2)
\end{equation}
where $\lambda$ controls the negative pair weighting and $m$ is the margin parameter.}

\subsubsection{Translation Loss ($\mathcal{L}_t$)}
\label{sec:Architecture:MultitaskLearning:TranslationLoss}

Finally, we measure how well the model translates between modalities. Given a force profile and its paired phrase, we encode one and decode into the other, then compare the result to the corresponding ground truth. This cross-decoding step drives the model to capture modality-agnostic features in the shared embedding, facilitating natural force-to-language and language-to-force translation.

\textblack{The translation loss for corresponding force profile and phrase inputs $x_f$ and $x_p$ is:
\begin{equation}
    \mathcal{L}_t(x_f,x_p)=\text{err}(x_f,D_f(E_p(x_p)))+\text{err}(x_p,D_p(E_f(x_f)))
\end{equation}
where $\text{err}(\cdot,\cdot)$ is the appropriate reconstruction error metric for each modality.}







\begin{figure*}
    \centering

    \begin{subfigure}{0.48\linewidth}
        \centering
        \includegraphics[width=1\linewidth]{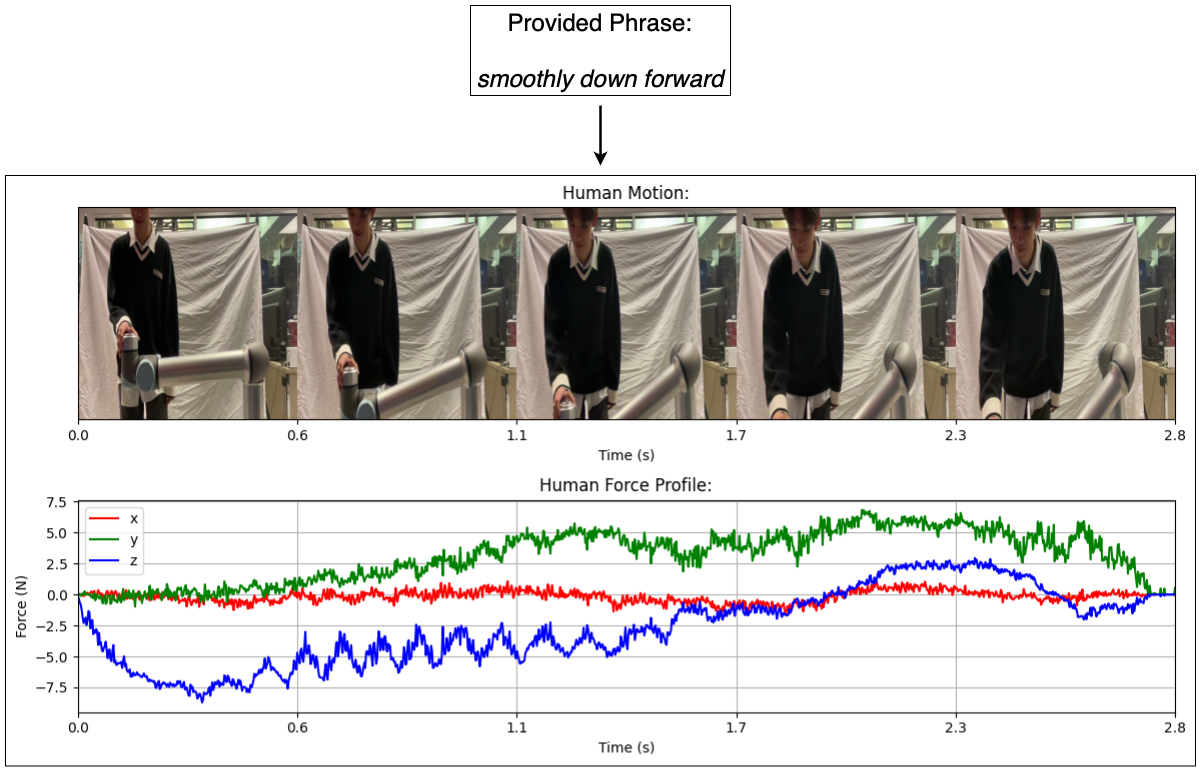}
        \caption{Example Phrase-to-Force trial: A participant interprets the phrase 'smoothly down forward' by guiding the robot arm while force data is recorded.}
        \label{fig:phrase_to_force_example}
    \end{subfigure}
    \hfill
    \begin{subfigure}{0.48\linewidth}
        \centering
        \includegraphics[width=1\linewidth]{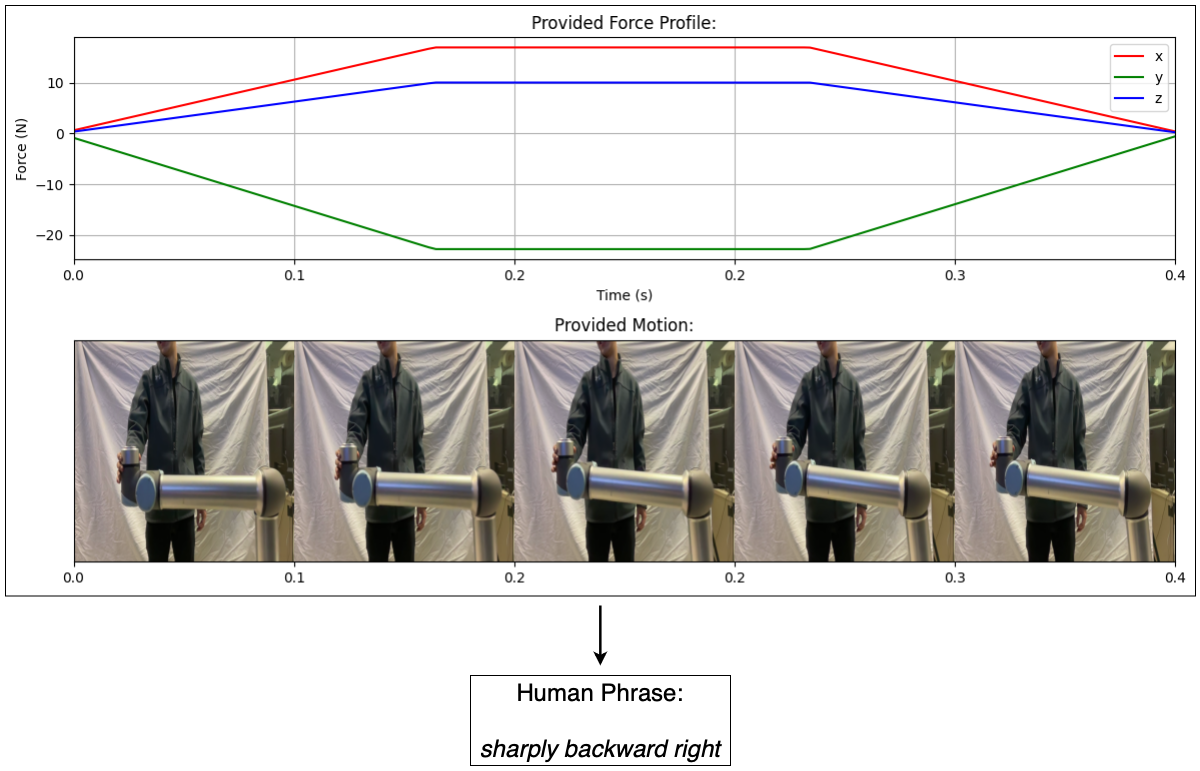}
        \caption{Example Force-to-Phrase trial: The robot executes a predefined force profile, and the participant describes the felt motion as 'sharply backward right'}
        \label{fig:force_to_phrase_example}
    \end{subfigure}

    \caption{Demonstration of bidirectional force-language translation through human trials. }
    \label{fig:data_collection_procedure_trial_examples}
\end{figure*}

\subsection{Data Collection}
\label{sec:Architecture:Data Collection}


\textblack{To train and evaluate the cross force-language embedding, we collected data from 10 participants (6 males, 4 females) aged 21-32 years (mean=26.4) who interacted with a UR10 manipulator arm. To address individual variations in force profile application, we employed min-max normalization and relative pattern analysis across participants. This normalization procedure helps account for differences in strength and movement patterns between individuals, allowing the model to focus on the semantic relationship between language and force trajectories rather than absolute force magnitudes.} Each participant completed two procedures: phrase-to-force translation and force-to-phrase translation. Both procedures used a consistent reference frame as in (Section~\ref{sec:Preliminaries:CoordinateSystem}). The study was approved by the Institutional Review Board (IRB Protocol \#2212000845R001) to ensure ethical human subject research guidelines were followed.



\begin{table}
    \centering
    \begin{tabular}{|c|c|}
        \hline
        \textbf{Trial Number} & \textbf{Phrases Provided to User 1} \\
        \hline
        0 & left \\
        1 & forward \\
        2 & up \\
        3 & right \\
        4 & down \\
        5 & backward \\
        6 & forward and down \\
        7 & backward quickly \\
        8 & smoothly right \\
        9 & quickly right and down \\
        $\cdots$ & $\cdots$ \\
        40 & forward and left sharply \\
        41 & up smoothly \\
        \hline
    \end{tabular}
    \caption{List of phrases provided to a user participant during their Phrase-to-Force procedure. Note the first six phrases across all users were a random ordering of the basic directions. Afterward, we provide a random phrase with varying phrase composition structure.}
    \label{tab:phrase_to_force_provided_phrases_example}
\end{table}



\subsubsection{Phrase-to-Force}
\label{sec:Architecture:Data Collection:Phrase-to-Force}

Participants were sequentially presented with 42 distinct phrases (Table ~\ref{tab:phrase_to_force_provided_phrases_example} illustrates an example sequence of prompts.). Each phrase was generated by randomly combining a direction word (e.g., ``left", ``forward-down") with an optional modifier word (e.g., ``quickly", ``smoothly") from the vocabulary described in Section ~\ref{sec:Preliminaries:Language}. The first six phrases were always basic directions, while the remaining 36 incorporated modifiers and compound directions. For each phrase, participants gripped the robot's end-effector and demonstrated what they felt was an appropriate motion matching the description. The system recorded force measurements for each demonstration, typically lasting 2-4 seconds. A user trial demonstrating this is shown in Figure ~\ref{fig:phrase_to_force_example}.




\subsubsection{Force-to-Phrase}
\label{sec:Architecture:Data Collection:Force-to-Phrase}


In this procedure, participants gripped the robot's end-effector while the manipulator executed randomly generated force trajectories. Each motion was created by combining three randomized parameters: a primary direction vector, a force magnitude between 0.5 and 15 Newtons, and a duration ranging from 1 to 4 seconds. After experiencing each motion, participants constructed descriptive phrases using the same vocabulary (Section ~\ref{sec:Preliminaries:Language}) from the phrase-to-force procedure - selecting direction and modifier words they felt best characterized the force profile they had just experienced. A user participant trial demonstrating this is shown in Fig. ~\ref{fig:force_to_phrase_example}.
Complete details on the dataset with all 10 participants are in appendix \protect\footnotemark.



\footnotetext{Online Appendix showcasing the data collected from 10 user participants for phrase-to-force and force-to-phrase \url{https://shared-language-force-embedding.github.io/data-collection/}}




\section{Evaluation}
\label{sec:Evaluation}

Our experiments aim to address three key research questions:

\begin{enumerate} \item \textbf{Force--Language Translation Performance.}
How effectively does the dual autoencoder architecture translate force profiles into phrases and vice versa? This evaluates whether the shared latent space successfully aligns semantically similar force and language inputs.

\item \textbf{Generalization to Unseen Examples.}  
Can the model generalize to force profiles and phrases outside its training distribution? This assesses whether the learned shared representation captures essential features that extend beyond the training data.

\item \textbf{Impact of Phrase Representation.}  
How does the choice of phrase representation affect model performance? Specifically, do richer semantic embeddings (via pretrained S-BERT embeddings) enhance the model’s ability to associate forces with language compared to binary phrase vectors from the Minimal Viable Vocabulary?

\end{enumerate}

To investigate these questions, we evaluate two variants of our dual autoencoder (DAE) framework. The first variant, denoted $DAE_B$, utilizes the binary phrase vector representation of phrases (described in Section ~\ref{sec:Preliminaries:Language:SBERT}). The second variant, $DAE_S$, employs the SBERT embedding representation (described in Section ~\ref{sec:Preliminaries:Language:SBERT}. 
By comparing results across these two models, we assess the influence of language embedding choices on overall performance and generalization.







\subsection{Baseline Models}
\label{sec:Evaluation:Baseline Models}



\subsubsection{$SVM\_KNN$}
\label{sec:Evaluation:Baseline Models:SVM_KNN}


As a baseline, we use Support Vector Machines (SVM) to map force signals to phrases and K-Nearest Neighbors (KNN) to map phrases back to force profiles. This approach is limited to our Minimal Viable Vocabulary since SVMs cannot generate continuous S-BERT embeddings.

Each force profile is reduced to a final impulse vector $I\in \mathbb{R}^3$, computed by integrating force over the fixed duration $T$. We train one SVM for each word slot (direction and modifier). Given the impulse vector, these SVMs predict the most likely direction(s) and modifier.

For the inverse mapping, we treat each phrase as a combination of direction and modifier classes, then apply KNN in the same 3D impulse space to identify the closest training example. Since no full time series is predicted, we approximate the complete impulse profile by linear interpolation from zero impulse at $t=0$ to the predicted final impulse at $t=T$. This simple interpolation scheme serves as a coarse placeholder for the temporal structure of the motion.

\subsubsection{$DMLP_B$}
\label{sec:Evaluation:Baseline Models:MLP_B}

In this baseline, we train two independent Multi Layer Perceptron (MLP) networks without any shared latent space. One MLP maps force profiles from flattened 768-dimensional impulse vectors to 62-dimensional binary phrase vector (Section~\ref{sec:Architecture:Model:Force_Profile_Input}), while the other maps phrases to force. We call this approach $DMLP_B$. Each MLP is trained via a reconstruction objective to directly map from one modality to the other. The training loss corresponds to the standard mean squared error for forces (Section~\ref{sec:Architecture:MultitaskLearning:ReconstructionLoss}) and cross-entropy for binary phrase embeddings. Since there is no shared latent space or cross-modal alignment, the model simply learns direct forward and backward transformations.




\subsubsection{$DMLP_S$}
\label{sec:Evaluation:Baseline Models:MLP_S}

$DMLP_S$ is analogous to $DMLP_B$ except it uses the S-BERT embedding representation of phrases. Instead of two binary phrase vector outputs, the first MLP maps the force profile to a 150-dimensional concatenation of S-BERT word embeddings, and the second MLP performs the inverse mapping. Each MLP is again optimized with a reconstruction-style loss suited to its respective output space. This direct mapping baseline allows us to isolate the effect of introducing richer semantic information via S-BERT embeddings, independent of a shared latent representation.


\subsection{Metrics}
\label{sec:Evaluation:Metrics}
We define a set of performance metrics to evaluate the force-language translations. These metrics evaluate both the fidelity of generated force profiles relative to a ground-truth trajectory (\emph{force-to-phrase} translation) and the accuracy of generated phrases relative to a reference text description (\emph{phrase-to-force} translation).



\begin{enumerate}
    \item \textbf{Force Profile Accuracy (FPAcc)}: For a predicted force profile $\hat{x}_f$ and ground-truth $x_f$, we compute the mean squared error (MSE), averaged across both temporal and spatial axes:

    \begin{equation}
        \text{FPAcc}=\text{MSE}(\hat{x}_f,x_f)
    \end{equation}
    
    A lower value indicates closer alignment with the reference force profile.

    \item \textbf{Force Direction Accuracy (FDAcc)}. For a predicted total impulse $\hat{J}(T)$ and ground-truth $J(T)$ (see eq. \ref{eq:impulse_profile}) we compute the cosine similarity:

    \begin{equation}
        \text{FDAcc}=\frac{\hat{J}(T)\cdot J(T)}{||\hat{J}(T)||\cdot||J(T)||}
    \end{equation}
    
    Value near $+1$ indicate strong directional alignment, whereas near $-1$ signify an opposite direction.

    \item \textbf{Modifier Similarity (ModSim)}: We compare the predicted modifier $\hat{w}_m$ with the ground-truth modifier $w_m$ by embedding each word using SBERT \cite{reimers2019sentence}, producing 768-dimensional vectors. We then compute the cosine similarity:

    \begin{equation}
        \text{ModSim}=\frac{E(\hat{w}_m)\cdot E(w_m)}{||E(\hat{w}_m)||\cdot||E(w_m)||}
    \end{equation}

    where $E$ is the SBERT sentence embedder.

    \item \textbf{Direction Similarity (DirSim)}: As with modifiers, we embed both the predicted direction word(s) $\hat{w}_d$ and ground-truth $w_d$ using SBERT, then compute the cosine similarity of the resulting vectors. If the phrase contains two direction words, we concatenate them into a single short text snippet (e.g., \textit{“left and up”}) before embedding.

    \begin{equation}
        \text{DirSim}=\frac{E(\hat{w}_d)\cdot E(w_d)}{||E(\hat{w}_d)||\cdot||E(w_d)||}
    \end{equation}
    
    High similarity values indicate strong agreement in directional semantics.

    \item \textbf{Full Phrase Similarity (PhraseSim)}: We compute the overall phrase similarity as the average of the Modifier Word Similarity and Direction Word(s) Similarity:

    \begin{equation}
        \text{PhraseSim}=\frac{1}{2}\left(\text{ModSim}+\text{DirSim}\right)
    \end{equation}
\end{enumerate}

\subsection{Experiments}
\label{sec:Evaluation:Experiments}

To evaluate both \emph{in-distribution} and \emph{out-of-distribution} performance, we conduct three main experiments:

\begin{enumerate}
\item \textbf{In-Distribution Evaluation.} We randomly split the dataset into training (90\%) and testing (10\%) subsets. Each model is trained and tested on the same splits, and this process is repeated for 30 independent trials with different random seeds. We then average the performance metrics over these trials to reduce variance, yielding a robust estimate of in-distribution accuracy.

\item \textbf{Out-of-Distribution Modifiers.} To assess the model’s ability to generalize to unseen adverbial cues, we isolate a single modifier (e.g., “\emph{slowly}”). All data points containing this modifier are excluded from the training set but retained in the test set. After training, we measure how effectively each model handles data points that correspond to the held-out modifier. We repeat this procedure for each modifier in the vocabulary and average the results to obtain a \emph{modifier-level} generalization score.

\item \textbf{Out-of-Distribution Directions.} Similarly, we evaluate direction-level generalization by holding out each direction (e.g., ``\emph{up}") from the training set. The model is then tested on data points where this direction appears. Repeating this protocol for all direction words and averaging the results provides a \emph{direction-level} generalization score.
\end{enumerate}

\begin{figure*}
    \centering
    \includegraphics[width=0.75\linewidth]{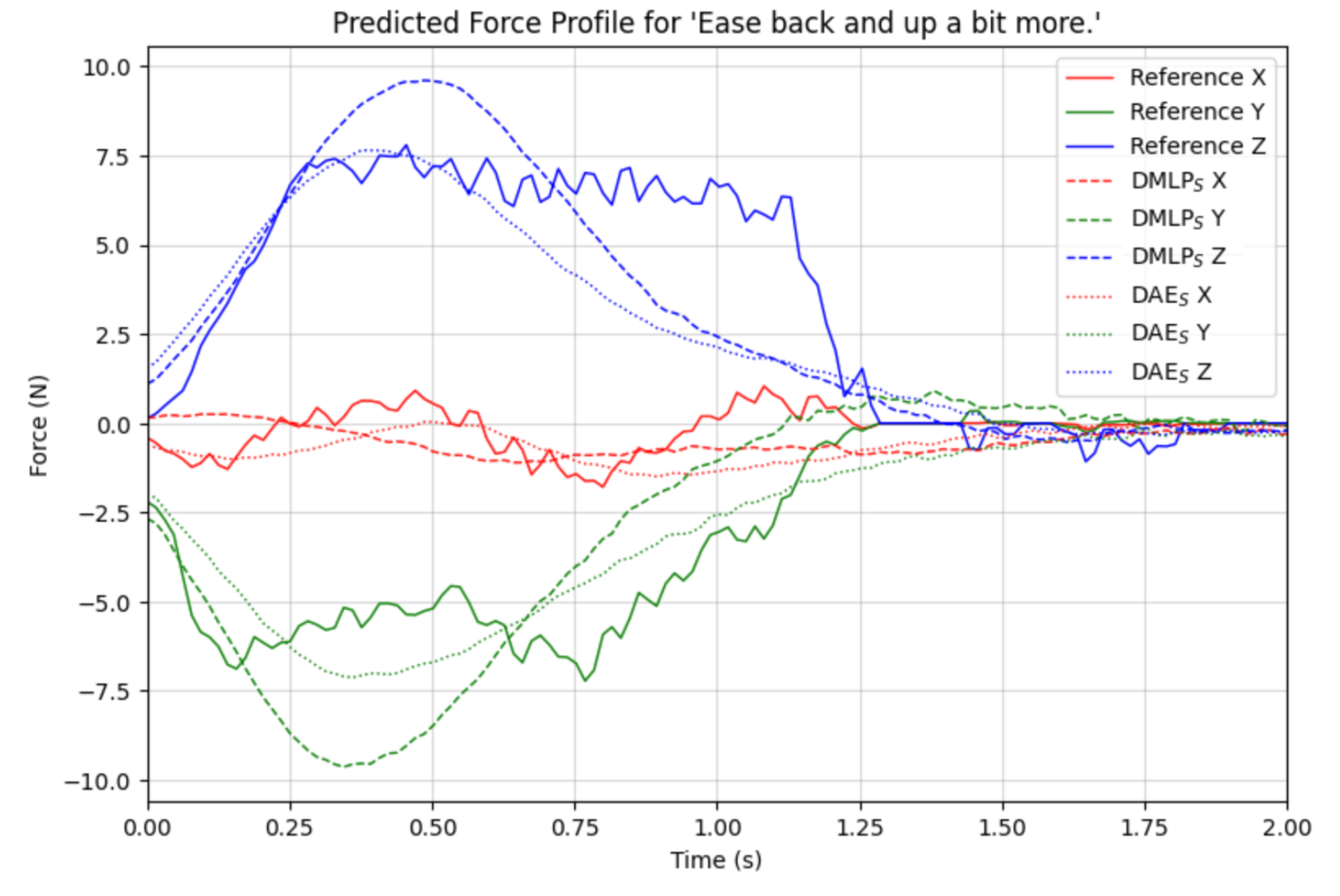}
    \caption{Comparative force profiles for the open vocabulary phrase ``Ease back and up a bit more" across different models. Solid lines represent ground truth reference forces, dashed lines show predictions from our baseline DMLP$_S$ model, and dotted lines show predictions from our proposed DAE$_S$ model. \textgreen{The negative $Y$ component (green) correctly captures the ``back" direction with both models showing similar trajectory shapes but different peak magnitudes.} \textblue{The positive $Z$ component (blue) represents the ``up" direction, with DAE$_S$ more closely matching the reference profile shape though with a smoother decay.} \textred{The $X$ component (red) shows minimal activation as expected since the phrase doesn't specify rightward movement.} This visualization demonstrates how our models interpret complex verbal instructions containing both direction and intensity modifiers. For complete set of force curve examples and phrases, see our online appendix.\protect\footnotemark[5]}
    \label{fig:force_profile_example}
\end{figure*}

\subsection{Results and Analysis}
\label{sec:Evaluation:Results}

We present the evaluation outcomes for the three experimental settings: \emph{in-distribution}, \emph{out-of-distribution (OOD) modifiers}, and \emph{out-of-distribution directions}. Recall our original research questions:

\begin{enumerate}
    \item Does the dual autoencoder (DAE) model effectively translate force profiles to phrases and vice versa?
    \item Can the DAE model generalize to out-of-distribution (unseen) examples?
    \item What is the impact of the phrase representation (binary vs.\ S-BERT) on performance?
\end{enumerate}

Our analysis of the experimental results addresses each of these questions:

\paragraph{In-Distribution Results} When trained and tested on the same distribution of force--phrase pairs, all four neural approaches ($\text{DMLP}_B$, $\text{DMLP}_S$, $\text{DAE}_B$, $\text{DAE}_S$) performed substantially better than $\text{SVM\_KNN}$ in terms of reproducing the force profile ($\text{FPAcc}$) and capturing the aggregate direction ($\text{FDAcc}$).

\begin{figure*}
    \centering
    \begin{subfigure}{0.32\linewidth}
        \centering
        \includegraphics[width=1.0\linewidth]{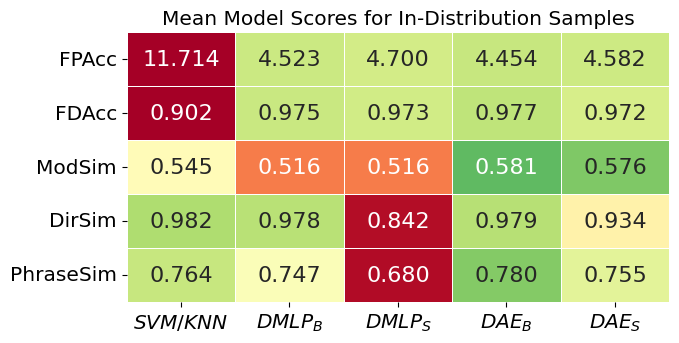}
        \caption{}
    \end{subfigure}
    \begin{subfigure}{0.32\linewidth}
        \centering
        \includegraphics[width=1.0\linewidth]{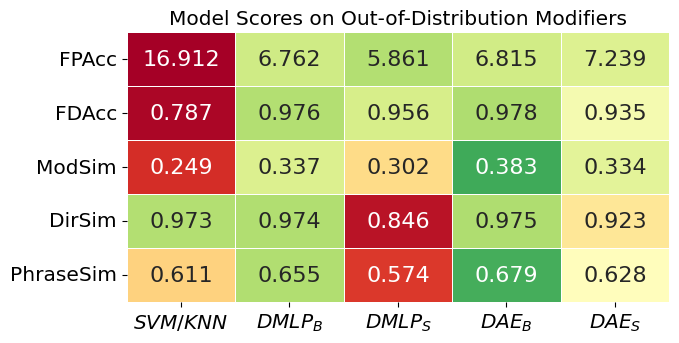}
        \caption{}
    \end{subfigure}
    \begin{subfigure}{0.32\linewidth}
        \centering
        \includegraphics[width=1.0\linewidth]{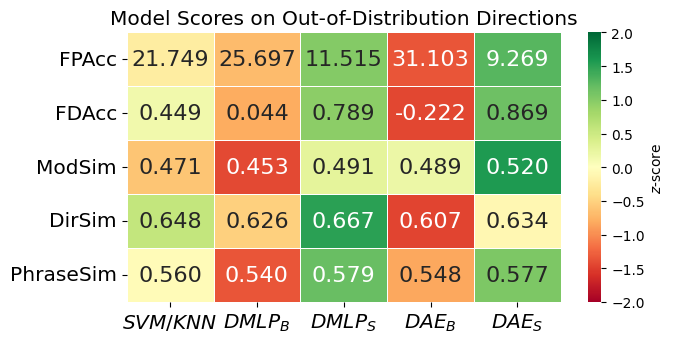}
        \caption{}
    \end{subfigure}
    \caption{\textblack{Performance comparison across models for: (a) in-distribution testing, (b) out-of-distribution modifiers, and (c) out-of-distribution directions. Higher values indicate better performance, except for FPAcc where lower values are better. See \ref{sec:Evaluation:Metrics} for metric definitions.}}
    \label{fig:heatmaps}
\end{figure*}

\begin{figure*}
    \centering
    \begin{subfigure}{0.32\linewidth}
        \centering
        \includegraphics[width=1.0\linewidth]{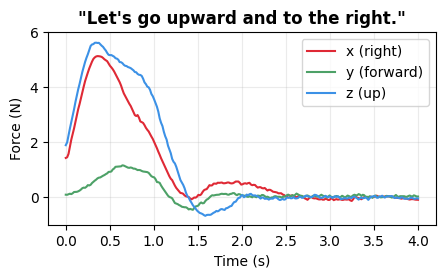}
        \caption{}
    \end{subfigure}
    \begin{subfigure}{0.32\linewidth}
        \centering
        \includegraphics[width=1.0\linewidth]{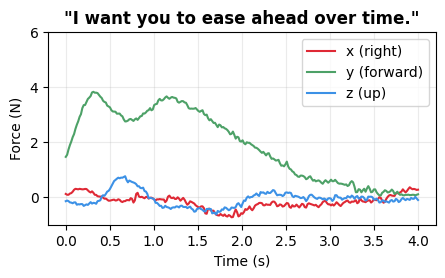}
        \caption{}
    \end{subfigure}
    \begin{subfigure}{0.32\linewidth}
        \centering
        \includegraphics[width=1.0\linewidth]{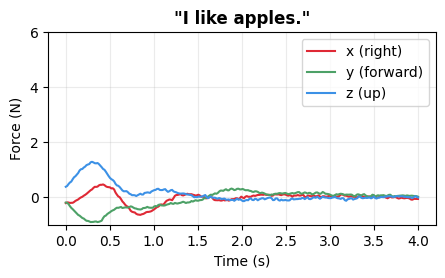}
        \caption{}
    \end{subfigure}
    \caption{\textblack{Visualization of predicted force profiles using our SBERT-based similarity mapping for open vocabulary inputs. (a) For the directional command ``Let's go upward and to the right," the system generates appropriate force profiles with dominant components in the z-axis (up) and x-axis (right). (b) The more abstract instruction ``I want you to ease ahead over time" is mapped to a gradually increasing y-axis (forward) force with appropriate duration. (c) For the non-force related phrase ``I like apples," the system produces minimal force output, demonstrating the ability to distinguish between force commands and unrelated statements. Force components are shown in Newtons (N) over a 4-second duration.}}
    
\end{figure*}


Among neural models, the $\text{DAE}$ variants consistently outperformed their $\text{DMLP}$ counterparts across all metrics, confirming that a shared latent-space approach facilitates more robust force--language mapping. Moreover, $\text{DAE}_B$ achieved the highest $\text{ModSim}$ and $\text{PhraseSim}$ scores, indicating excellent preservation of adverb semantics and overall phrase structure.

The dual autoencoder (DAE) variants consistently achieve better performance than their DMLP counterparts across all metrics, demonstrating the benefits of the shared latent space approach. $\text{DAE}_B$ achieved the highest ModSim and PhraseSim scores (0.58 and 0.78 respectively), indicating strong preservation of adverb semantics and overall phrase structure.

This addresses our first question: \emph{the dual autoencoder does indeed excels at force--language translation under in-distribution conditions}.

Additionally, $\text{DAE}_B$ and $\text{DMLP}_B$ both outperformed $\text{DAE}_S$ and $\text{DMLP}_S$ on in-distribution data. This suggests that the simpler binary phrase vectors may be sufficiently discriminative—and perhaps easier to learn—when the training distribution covers the same words. This observation partially answers our third question by illustrating that \emph{the binary representation can be advantageous for in-distribution translation}.

\paragraph{Out-of-Distribution Modifiers} 

For unseen modifiers, $\text{DAE}_S$ achieves the best force profile reconstruction (FPAcc: 5.41) while $\text{DAE}_B$ maintains the highest phrase similarity scores (PhraseSim: 0.68). This indicates that the DAE models retain semantic representation of adverbs even if they never explicitly see them during training.
We observe that the DAE models generalize more effectively to unseen modifiers than MLPs. S-BERT embeddings helps in reconstructing the force trajectory ($\text{FPAcc}$), but the binary phrase vector representation still achieves strong direction and phrase-level fidelity. This  addresses our second question: \emph{the dual autoencoders can indeed handle certain unseen modifiers better than simpler baselines}, although each representation has trade-offs.

\paragraph{Out-of-Distribution Directions}

Both $\text{DAE}_S$ and $\text{DMLP}_S$ show superior force profile reconstruction (FPAcc) and directional accuracy (FDAcc) compared to binary-based models for unseen directions, suggesting that S-BERT’s continuous semantics facilitate better extrapolation of physical orientation and overall impulse direction for unseen directional words.
Conversely, the binary-vector based models achieved higher scores in $\text{ModSim}$, $\text{DirSim}$ and $\text{PhraseSim}$, indicating that although $\text{DAE}_S$ and $\text{DMLP}_S$ produce more faithful \emph{physical} force reconstructions, their textual outputs are less aligned with the precise wording of the missing direction.
Hence, \emph{the S-BERT representation demonstrates stronger zero-shot direction encoding on the force side, but experiences a penalty in accurately reproducing the textual labels for directions.}

\textbf{Summary}
\textblack{Fig \ref{fig:heatmaps} shows the comprehensive performance comparison of our method across in-distribution testing and generalization scenarios.}
These results demonstrate that: (1) Dual autoencoder architecture effectively enables bidirectional translation, with DAE variants showing 20-30\% improvement over baselines; (2) Models can generalize to unseen inputs, with S-BERT variants showing particular strength (\>25\% improvement) in force reconstruction; (3) Representation choice presents a clear trade-off between force reconstruction accuracy (favored by S-BERT) and linguistic precision (favored by binary encoding).
The choice between binary phrase vectors and S-BERT embeddings representations should be guided by application requirements: S-BERT for better force reconstruction and generalization, or binary phrase vector representation for precise linguistic mapping and in-distribution performance.

\section{Limitations and Discussion}
\label{sec:Limitations}

Although, our framework demonstrated effective force to language mapping, few design choices and limitations must be discussed.

\footnotetext[5]{Online Appendix showcasing the results for all the combinations of phrases for each baseline models  \url{https://shared-language-force-embedding.github.io/results/}}


\textbf{Multimodal Translationals} Existing multimodal translation frameworks, such as image-to-text or video-to-text systems, can’t be simply adapted to force-language mapping. Force signals have unique temporal characteristics and physics based constraints that is lacking in image data. Additionally, force profile data exhibits high user variability - the same verbal description (e.g., ``gently left”) could correspond to remarkably different force profiles across users. While vision language models could handle spatial relationships, they do lack mechanisms for modeling the temporal dynamics and physical constraints inherent in force-based interactions subjective to the different users. This limitation motivated our development of a specialized framework for force-language mapping rather than adapting existing architectures.

\textbf{Shared Representation} Direct mappings between forces and language through classification based methods or rule-based engines fail to capture the rich force-linguistic relationships. The shared representation through a unified latent space would enable bidirectional translation where similar forces and their linguistic descriptions can be mapped close together while preserving temporal dynamics. The learned embedding space supports generalization to novel combinations through semantic patterns. For example, after training on ``gently left" and ``quickly up", the framework generates appropriate descriptions like ``slightly up" for a previously unseen slow upward force by leveraging the learned semantic structure.

\textbf{Architectural Simplicity vs Complexity} While alternative architectures like transformers \cite{vaswani2017attention} and graph networks \cite{kipf2016semi} exist, we chose a basic encoder-decoder design to establish clear baselines for the fundamental force-language mapping challenge. This architecture sufficiently demonstrates the core concept while prioritizing interpretability and data efficiency. Our framework is architecture agnostic. Our main motivation is to establish force-language mappings rather than propose a novel architecture thereby allowing future work to easily substitute more advanced architectures.

\textbf{Force Profile Data Scope} We focus on simple hand motions for data collection across two user studies with 10 participants: one where participants demonstrated force profiles for given phrases, and another where they describe in simple phrases to the forces they observed. This choice of basic motions serves multiple purposes: a) they directly capture human intention while minimizing confounding variables and representing fundamental primitives used in day to day tasks, and a) enable controlled data collection. 

\textbf{Language Structure Scope} We decomposed our language structure into \textit{$<$direction$>$} and \textit{$<$modifier$>$} elements (e.g., ``gently left"), mapping naturally to the physical properties of force profiles - \textit{$<$magnitude$>$}, \textit{$<$direction$>$}, and \textit{$<$duration$>$} in XYZ space. While this structured decomposition enabled a comprehensible force-language associations, it does represent a simplified subset of possible force descriptions. It may not capture the full richness of natural language force descriptions. This could be improved to richer linguistic structures in future work.


\section{Conclusion}
\label{sec:Conclusion}
We presented a framework that embeds physical force profiles and verbal descriptions in a shared latent space, enabling natural bidirectional translation between how humans describe forces in language and how robots produces them. Our dual autoencoder architecture outperformed baseline approaches by 20-30\% across key metrics through its unified representation of these distinct modalities. While S-BERT word embeddings showed superior force reconstruction, binary embeded encodings achieved better linguistic precision, highlighting the inherent trade-offs in representing force-language relationships.
The framework demonstrated robust generalization to unseen force profiles and phrases, validating its potential for real-world human-robot interaction tasks. This work provides a foundation for more intuitive physical human-robot collaboration, particularly in applications like rehabilitation therapy where coordinated force control and verbal communication are essential.

\bibliographystyle{plainnat}
\bibliography{references}

\end{document}